\pdfoutput=1

\documentclass[11pt]{article}
\usepackage[table]{xcolor}
\usepackage[final]{coling}

\usepackage{times}
\usepackage{latexsym}
\usepackage[draft]{changes}
\usepackage[T1]{fontenc}

\usepackage[utf8]{inputenc}

\usepackage{microtype}

\usepackage{inconsolata}

\usepackage{graphicx}
\usepackage{amsmath}
\usepackage{tabularx}
\usepackage{array}
\usepackage{cancel}
\usepackage{multirow}
\newcolumntype{Y}{>{\centering\arraybackslash}X}

\title{MLD-EA: Check and Complete Narrative Coherence by Introducing Emotions and Actions}

\author{Jinming Zhang \ Yunfei Long \\
    University of Essex \\
  \texttt{jz22273@essex.ac.uk} \ \texttt{yl20051@essex.ac.uk} \\}

\begin{document}
\maketitle
\begin{abstract}
Narrative understanding and story generation are critical challenges in natural language processing (NLP), with much of the existing research focused on summarization and question-answering tasks. While previous studies have explored predicting plot endings and generating extended narratives, they often neglect the logical coherence within stories, leaving a significant gap in the field. To address this issue, we introduce the \textbf{M}issing \textbf{L}ogic \textbf{D}etector by \textbf{E}motion and \textbf{A}ction (\textbf{MLD-EA}) model, which leverages large language models (LLMs) to identify narrative gaps and generate coherent sentences that integrate seamlessly with the story's emotional and logical flow. The experimental results demonstrate that the MLD-EA model enhances narrative understanding and story generation, highlighting LLMs' potential as effective logic checkers in story writing with logical coherence and emotional consistency. This work fills a gap in NLP research and advances border goals of creating more sophisticated and reliable story-generation systems.
\end{abstract}

\section{Introduction} \label{Introduction}
Narrative understanding and story generation have been a compelling challenge in Natural Language Processing (NLP) for a long. They evolved from early rule-based systems with limited creativity to sophisticated models that generate rich, engaging narratives \cite{mooney1985learning, fan2018hierarchical}. Introducing Transformer \cite{vaswani2017attention} models like BART \cite{lewis2020bart} and large language models (LLMs) like ChatGPT \cite{openai2024chatgpt} revolutionized this task by utilizing advanced architectures to capture in-detailed dependencies.

Many previous studies have focused on tasks like summarizing \cite{awasthi2021natural, jin2024comprehensive}, sentiment analysis \cite{lu2023sentiment, zhao2025leveraging, lu2025multimodal} and question-answering (QA) \cite{zhuang2024toolqa, huang2024prompting}. While previous story generation research often centered on predicting plot endings or crafting long narratives \cite{guan2020knowledge, li2022diffusion}. However, in general, story writing frequently needs to pay more attention to maintaining logical coherence \cite{oatley2002emotions, currie2004narrative}. 

Not surprisingly, some recent works lead LLMs to maintain narrative coherence in different ways with effective results \cite{zhao2023more, wang2023improving}. However, most of those works focus on continuously writing coherency stories by LLMs \cite{guan2021long}. There is still a gap in detecting the logical coherence in the narratives. 

To address this gap, our approach focuses on the observable actions of characters rather than delving into their deeper motivations. This choice stems from the understanding that actual actions have a more immediate and direct impact on emotions, and conversely, emotions are often the driving force behind tangible actions \cite{zhu2002emotion, doring2003explaining}. The James-Lange theory of emotion in psychology posits that physiological responses to a situation—such as a racing heart or clenched fists—occur first and then lead to the subjective experience of emotion \cite{cannon1927james}. This suggests that an observable action (like a person slamming a door) can directly trigger an emotional response (such as anger or frustration). Similarly, the cognitive-behavioral theory emphasizes that behaviors (actions) and emotions are closely linked, where a behavior change can directly influence emotional states, and vice versa \cite{maslow1943theory, eisenberg2014altruistic, leahy2022cognitive}.

By prioritizing the direct interplay between observable actions and emotions, we aim to capture the essence of narrative logic in a way that reflects these well-documented psychological principles \cite{carver2000action}. This approach is supported by extensive psychological studies that emphasize the strong correlation between actions and emotional responses, such as how consistent patterns of behavior can shape long-term emotional states, as seen in theories of learned helplessness or social learning \cite{bandura1977social}.

In this study, we introduce the \textbf{M}issing \textbf{L}ogic \textbf{D}etector by \textbf{E}motion and \textbf{A}ction (MLD-EA), a LLM-based model designed to identify gaps in narrative logic and generate missing plot elements that are coherent both logically and emotionally. By incorporating the relationship between actions and emotions, MLD-EA aims to enhance the logical structure of narratives. Experimental results demonstrate that our models can produce more believable and emotionally coherent stories by aligning narrative generation with these psychological insights. Our model improves narrative understanding and story generation, underscoring the potential of LLMs as story generators and powerful logic checkers in the creative process. 

The main contributions of our work can be briefly summarized as follows: \textbf{1)}, We propose a novel task of narrative logic detection. 
\textbf{2)}, By grounding our model in cognitive-behavioral theories, we highlight how emotions directly interact with actions, leading to better narrative understanding and generation.
\textbf{3)}, Experiments have shown that our MLD-EA model has achieved superior results in most aspects, including narrative logic checking with involved characters' emotions and actions and missing plot completeness. Also, we demonstrate the importance of behavior and emotion in story logic detection and generation.

Leveraging this interaction between actions and emotions to assess and generate story logic more efficiently and accurately mirrors the natural cause-and-effect relationships in human behavior. 

\begin{figure}[h]
    \centering
    \includegraphics[width=\columnwidth]{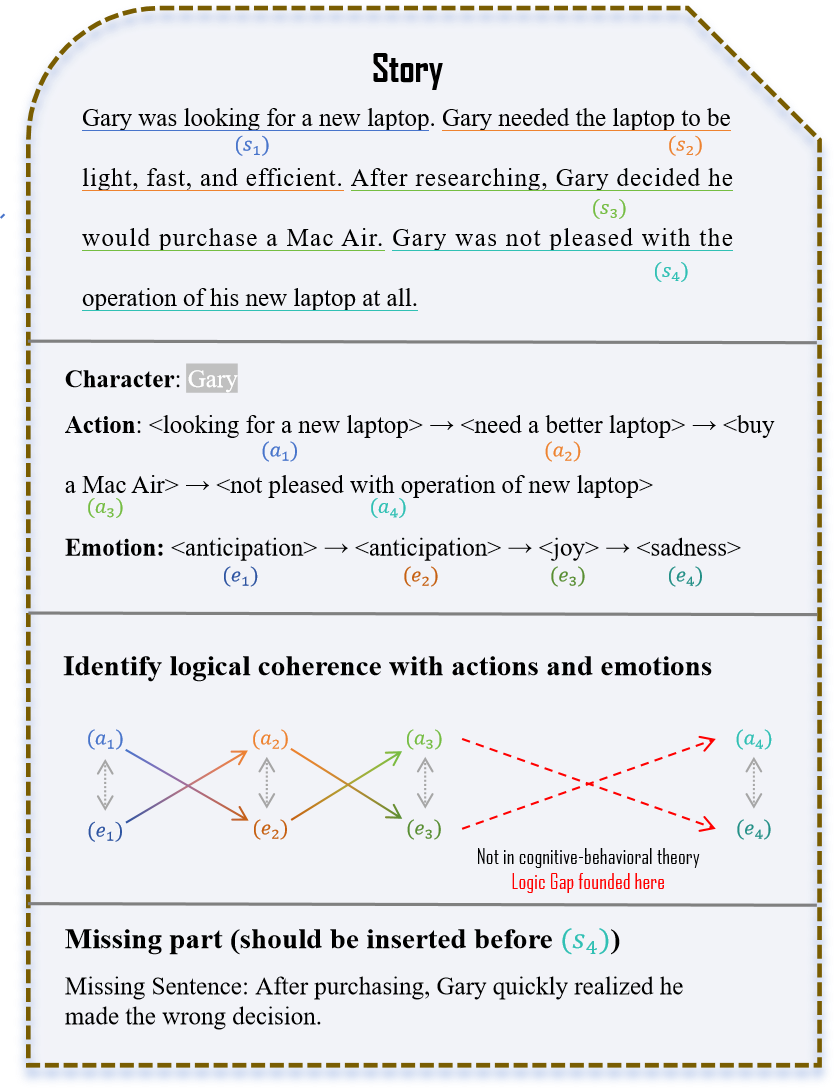}
    \caption{A task example. "Identify logical coherence with actions and emotions" is checking the logical coherence guided by the cognitive-behavioral theory.}
    \label{fig:Task Example}
\end{figure}

\section{Related Works} \label{Related works}
Several innovative approaches have been developed to enhance AI-generated narratives' logical coherence, emotional depth in narrative understanding, and story generation within NLP. \citet{paul2021coins} framework introduces a recursive inference strategy that dynamically generates contextualized rules to guide narrative completion, focusing on maintaining coherence and logical flow throughout the story. Similarly, the CHAE model \cite{wang2022chae} offers fine-grained control over narrative elements, creating customized stories with specific characters, actions, and emotions, enhancing the personalization and richness of the narratives. Similarly, the COMMA \cite{xie2022comma} explores the relationships among motivations, emotions, and actions, providing a cognitive framework that deepens the understanding of narrative construction by modeling these interrelated factors. However, these traditional models often struggle to consistently integrate actions and emotions to maintain logical coherence throughout the entire narrative, leading to disjointed or emotionally inconsistent storylines when handling more complex plots \cite{kambhampati2024llms}. Additionally, they may lack the flexibility to dynamically understand nuanced shifts in a character's behavior or emotional progression.

Exploring LLMs, cognitive frameworks, and hybrid planning strategies has paved the way for more engaging and human-like stories. \citet{alvarez2023chatgpt} used ChatGPT in interpreting narrative structures, which further extends the potential for generating stories based on predefined structures, offering new methods for narrative development. Notably, approaches such as iterative prompting-based planning for suspenseful story generation \cite{xie2024creating}, the combination of symbolic planning with neural models \cite{farrell2024planning}, and the SWAG method \cite{patel2024swag}, which utilizes action guidance in storytelling, have significantly improved the quality and engagement of AI-generated narratives. Additionally, comprehensive evaluations like "The Next Chapter" \cite{xie2023next} and knowledge-enhanced pre-training models \cite{guan2020knowledge} have shown that LLMs can produce stories of high quality, sometimes approaching the level of human authors. LLMs often struggle to maintain consistent plots on generation, but they cannot check their generated stories by themselves \cite{huang2024large}. In our approach, MLD-EA is able to find such logical loopholes by introducing the interaction between emotions and actions to keep stories coherent.

\begin{figure*}[h]
    \centering
    \includegraphics[width=\linewidth]{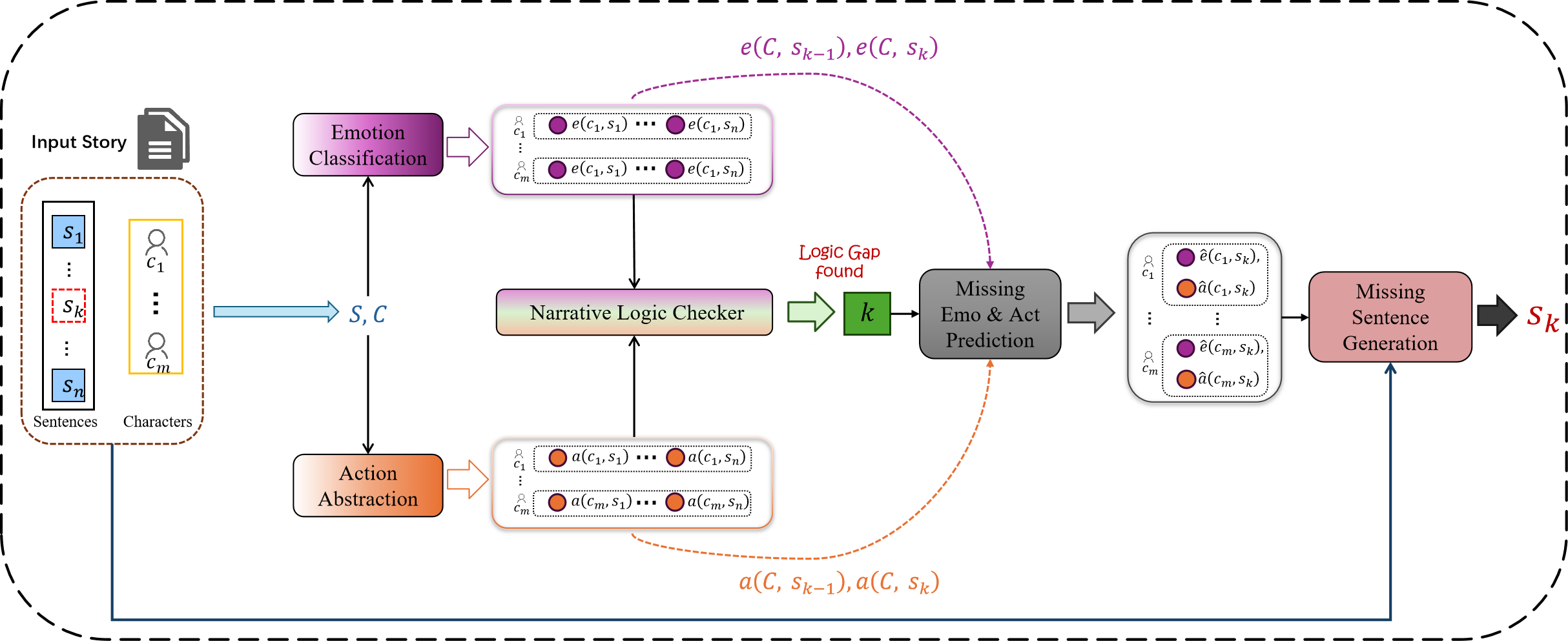}
    \caption{MLD-EA model overview. Each Input Story contains $n$ sentences and $m$ characters, which have a missing sentence $s_k$ before index $k$. $e(c,s)$ and $a(c,s)$ denote the character's emotion and action in the sentence, respectively; $\hat{e}$ and $\hat{a}$ denotes the predicted emotion and action.}
    \label{fig:MLD_EA pipline}
\end{figure*}

\section{Problem Definition} \label{Problem Definition}
The primary goal of MLD-EA is to identify whether the input story is logically completed, as Figure \ref{fig:Task Example} shows. We divided the model into four main sub-tasks: \textbf{1)}, abstracting characters' actions. \textbf{2)}, classifying their emotions for each sentence. \textbf{3)}, then locate the logical loopholes of the narrative in which the missing part should be inserted. \textbf{4)}, we complete the tale consistently by predicting the characters' actions and emotions. Thereby preserving the narrative's overall coherence and logical structure. The tasks are defined as follows: 

For any input $n$ sentences story ($S = s_1, \cdots, s_n $) with $m$ characters appeared in this story ($C=c_1,\cdots,c_m$), MLD-EA abstract characters' actions $a$ and classify their emotions $e$ for each sentence, denoted as $\{ (c, s) \rightarrow (a(c, s), e(c, s)) \mid c \in C, s \in S \}$, where $a(c, s)$ represents the action of character $c$ in sentence $s$ and $e(c, s)$ represents the emotion of character $c$ in sentence $s$. 

Sequently, given the story and characters' actions and emotions, MLD-EA will use the provided information to review the story and find inconsistencies. Notably, our task is to find the logic gap in the inner story. We suppose the start and end of the story are always complete. The process involves identifying points where the characters' actions or emotions exhibit abrupt changes that the preceding context cannot logically explain. After that, MLD-EA outputs the index $k$ which the missing part should be inserted before it:
\begin{equation}
    k =
    \begin{cases}
        1 < k < n & \textit{if there is a } \\
         & \textit{missing sentence} \\
        -1 & \textit{otherwise}.
    \end{cases}
\end{equation}

Formally, if MLD-EA identifies a logic gap before a specific place $k$ in the story, it proceeds by predicting the most likely actions $\hat{a}(c, s_k)$ and emotions $\hat{e}(c, s_k)$ by using the sequence of preceding ($ \{ a(c, s_{k-1}), e(c, s_{k-1}) \}$) and succeeding actions and emotions ($ \{ a(c, s_k), e(c, s_k) \}$). Then MLD-EA estimates the most coherent missing sentence $s_k$ according to $\hat{a}(c, s_k)$ and $\hat{e}(c, s_k)$. 


\section{Methodology} \label{Methodology}
In this section, we will provide a detailed methodology for each module within our MLD-EA model. The model architecture is shown in Figure \ref{fig:MLD_EA pipline}.

\subsection{Action Abstraction} \label{Action Abstraction}
The action abstraction module is designed to extract and abstract actions performed by characters in a given sentence, playing a crucial role in analyzing narrative structures and identifying logic gaps. The process begins with the model receiving a sentence \( s \), a list of characters $C = \{c_1, c_2, \dots, c_m\}$, and the story's context $S$ for reference. 

Guided by prompt engineering (details in Appendix \ref{Prompts Engineering}), MLD-EA processes each sentence to identify and represent the actions performed by the characters as flowing:

For each character $c$ in the characters list $C$, the model outputs an action in the following format: $\texttt{<c>Action(Target, Object)</c>}$, where $c$ represents the character acting; $\texttt{Action}$ denotes the action the character performs; $\texttt{Target}$ is the target of the action (who or what the action is directed towards); $\texttt{Object}$ specifies any object associated with the action (if applicable).
If a character $c$ does not perform any action in the sentence $s$, the model needs to output: $\texttt{<c>None</c>}$.

\subsection{Emotion Classification} \label{Emotion Classification}
The emotion classification module in the MLD-EA categorizes characters' emotions based on given sentences. This classification is based on eight basic emotion types from Plutchik's model \cite{plutchik2001nature} —\textit{joy, trust, fear, surprise, sadness, disgust, anger, and anticipation}—plus an additional "\textit{none}" category for cases where no emotion is detected.

Before classifying emotions, the model first checks whether each character $c$ in the list $C = \{c_1, c_2, \dots, c_m\}$ is affected by the events described in each sentence $s$. If the model determines that a character $c$ is not affected, the emotion for that character is classified as \texttt{none}. In addition to the emotion classification, the model also outputs whether or not each character is affected by the sentence.

The model's output for each character $c$ includes the result of the '\textit{affected}' and the emotion classification in $\texttt{<c>(Affected, $e(c,s))$</c>}$, where $\texttt{Affected}$ is a boolean value indicating whether the character $c$ is affected by  any event in the sentence $s$ and $e(c, s)$ represents the emotion associated with the character in sentence $s$, where $e \in \{ \texttt{joy}, \texttt{trust}, \texttt{fear}, \texttt{surprise}, \texttt{sadness}, \texttt{disgust},\\ \texttt{anger}, \texttt{anticipation}, \texttt{none} \}$.


\subsection{Narrative Logic Checker By Characters' Emotion and Action} \label{Missing Sentence Index Prediction}
The narrative logic checker component focuses on detecting potential gaps in the narrative by analyzing the relationship between characters' actions and emotions. This process is grounded in the outputs from the previous modules: action abstraction and emotion classification. The prediction is based on detecting disruptions or inconsistencies in each character's expected flow of actions and emotions.

Several key principles in behavior research  \cite{cannon1927james, zhu2002emotion} guide this process: 1), emotions often drive actions. 2), actions can influence subsequent emotions. 3), and some actions directly reflect the character's current emotional state, and vice versa. 



MLD-EA then predicts the missing sentence index $k$, which is determined by evaluating the continuity and logical consistency of the sequences with the interaction of characters' actions and emotions:
\begin{equation}
    \label{eq:emo and act}
    \begin{aligned}
        (E, A) =  \sum_{s \in S, c \in C} \left[ e(c, s), a(c, s) \right],
    \end{aligned}
\end{equation}
\begin{equation}
    \label{eq:missing index prediction}
    \begin{aligned}
        k = & \text{Inf}_{Index} \left[ (S \oplus (E, A)), C \right],
    \end{aligned}
\end{equation}
 where $\text{Inf}_{Index}$ represent the model inference of missing sentence index prediction. A significant deviation from expected values suggests a missing sentence, and $k$ identifies the position where this sentence should be inserted.

\subsection{Action/ Emotion prediction and sentence generation}
Following the identification of the missing sentence index by analyzing characters' actions and emotions, the next crucial step in the MLD-EA framework is to predict the actions and emotions of the missing sentence and subsequently generate the sentence. This process is essential to ensure the narrative remains coherent and logically consistent. The focus here is on the immediate context surrounding the predicted index. By examining the sequences of preceding actions and emotions and succeeding actions and emotions, the model estimates the most coherent actions \( \hat{a}(c, s_k) \) and emotions \( \hat{e}(c, s_k) \) for the missing sentence \( s_k \):
\begin{equation}
    \label{eq:act and emo prediction}
    \begin{aligned}
        \left[\hat{a}, \hat{e}\right] = 
        \text{Inf}_{eap}
        & \left[(a(c, s_{k-1}), e(c, s_{k-1})\right),  \\
        & \left(a(c, s_{k}), e(c, s_{k})) \right],      
    \end{aligned}
\end{equation}
where $\text{Inf}_{eap}$ means the model inference of emotion and action prediction for the missing sentence. Once these predictions are made, the model generates a sentence to fill the identified gap:
\begin{equation}
    \label{eq:sentence genration}
    s_k = \text{Inf}_{gen}(S, C, k, (\hat{a}, \hat{e} )),
\end{equation}
where $\text{Inf}_{gen}$ is a zero-shot inferring. This generated sentence encapsulates the character's possible emotion and action, thereby maintaining the story's coherence and flow and completing the narrative.

\section{Experiment} \label{Experiment}
\subsection{Data}
We use the Story Commonsense dataset for our task, which contains 4853 five-sentence stories with labeled emotions and motivation for characters \cite{rashkin2018modeling}. 
We only take the stories with labeled emotion because the labeled motivations are based on Maslow's needs \cite{maslow1943theory} and Reiss' motives \cite{reiss2004multifaceted} theory, which are focused on the deeper motivation, not actual actions. By excluding motivations, which are more abstract and theoretical in nature, the analysis remains more grounded in observable narrative events, avoiding complexities that may not directly influence the characters' visible actions. This also ensures that the model can better focus on the emotional states that drive the characters’ responses, making it easier to align predictions with surface-level events in the story.
We then divided the data into 8:1:1 for training, validation, and testing. 

To follow the task of emotion classification in section \ref{Emotion Classification} and the task of narrative logic checker in section \ref{Missing Sentence Index Prediction}, we consolidate the characters' emotions into a single tag by selecting the one with the highest confidence, as determined by three annotators in the original dataset. The details of choosing the missing sentence are in Appendix \ref{Chosen Missing Sentence}.

\subsection{Selected Baselines}
We compare MLD-EA with the following baselines trained by different strategies and datasets:

\textbf{Llama3-8B-Instruct} \cite{llama3modelcard}: Meta's Llama3-8B-Instruct model is a cutting-edge LLM renowned for its exceptional ability to follow instructions meticulously. It is adept at crafting stories that are not only imaginative but also adhere to logical structures and factual integrity. 

\textbf{Gemma2-2B-it} \cite{gemma_2024}: Gemma2-2B-it is a nimble and efficient model that packs a punch regarding text generation capabilities from Google. Despite its smaller size than some of its peers, it demonstrates remarkable skill in spinning engaging stories that captivate audiences. 

\textbf{Gemma2-9B-it} \cite{gemma_2024}: Gemma2-9B-it is a larger version of Gemma2-2B-it. With a more vast dataset and bigger model size, it generates intricate and vivid stories rich in detail and depth. 

We selected these particular models as baselines for several key reasons: 1) To the best of our knowledge, no prior research has focused on identifying logical gaps or inconsistencies at the sentence level within stories. This novel focus makes it difficult to directly compare our approach to existing studies. 2) While previous works on story generation have primarily relied on pre-trained models such as BERT and GPT-2 \cite{wang2022chae, paul2021coins}, our study specifically aims to evaluate the capabilities of newer LLMs. The baselines we selected models are all modern LLMs known for their advanced narrative understanding abilities. These models are particularly well-suited for complex tasks related to narrative. 3) We intentionally included models of different sizes and architectures to provide a comprehensive evaluation. This range allows us to compare varying complex models to understand how size and dataset diversity impact logical story generation.

\subsection{Implement Setups}
MLD-EA is built based on Llama3-8B-Instruct \cite{llama3modelcard} using the Huggingface's libraries\footnote{\url{https://huggingface.co/docs}} \cite{wolf2019huggingface} and use Llama-Factory \cite{zheng2024llamafactory} for supervised fine-tuning \cite{gunel2020supervised}. We use LoRA \cite{hu2021lora} to fine-tune our model. Please see Appendix \ref{hyper-parameters of fine-tune for MLD-EA} for hyper-parameters details and Appendix \ref{Prompts Engineering} for prompts technics we used and prompt templates.


We compute the micro-averaged result of all baselines by the same zero-shot \cite{wei2021finetuned}, one-shot, and few-shot \cite{brown2020language} prompts with original input labels from the dataset. All experiments run on two RTX 4090 24GB GPUs.

\subsection{Evaluation Metrics}
We use the following metrics to evaluate MLD-EA performance on the different sub-tasks: 

(1) Both \textbf{BLEU-1,2} \cite{papineni2002bleu} and \textbf{ROUGE-L} \cite{lin2004rouge} are used for evaluating the action abstraction task. 

(2) We compute the micro-average \textbf{Precision}, \textbf{Recall}, and \textbf{F1} score for each tag to show the accuracy of emotion classification.

(3) The micro-average \textbf{Precision}, \textbf{Recall}, and \textbf{F1} score are also applied to evaluate the accuracy of the narrative logic checker on each candidate place.

(4) For final generation task based on predicted emotions and actions, we use \textbf{BLEU-1,2,4}, \textbf{ROUGE-1,2,L} and \textbf{BERTScore}\footnote{The BERTScore evaluation model is from Hugging Face: \url{https://huggingface.co/spaces/evaluate-metric/bertscore}. We used the missing sentence from the original story as the reference and the model-generated sentence as predictions to compute their similarity.} \cite{zhang2019bertscore} to measure the similarity of candidate sentences and reference sentences. Furthermore, a \textbf{Valence-Arousal-Dominance (VAD)} model \cite{warriner2013norms} is used in psychology to describe and measure human emotions. These three dimensions are often used to provide a more comprehensive understanding of emotional states, as they capture different aspects of how emotions are experienced and expressed. We use a developed VAD model \cite{plisiecki2024extrapolation} to model the gap between candidate sentences and reference sentences. Also, \citet{plisiecki2024extrapolation} add Age of Acquisition (AoA) and Concreteness as important features in their VAD. AoA refers to the age at which a person learns a particular word or concept. 
Concreteness measures how tangible or perceptible through the senses a word is. 






\begin{table}[h] 
   \centering
    \footnotesize
        \begin{tabularx}{\linewidth}{Xccc}
            \hline\hline
            \textbf{}\rule{0pt}{2ex}       & \textbf{BLEU-1}     & \textbf{BLEU-2}     & \textbf{ROUGE-L} \\
            \hline
            T2Act2T\rule{0pt}{2ex}         & 40.94               & 34.82               & 53.67             \\
            \hline\hline
        \end{tabularx}
    \caption{Result of Action Abstraction}
    \label{tab:Result of Action Abstraction}
\end{table}

\begin{table}[h]
\centering
\footnotesize
    \begin{tabularx}{\linewidth}{Xccc}
        \hline\hline
        \textbf{Model}\rule{0pt}{2ex}      & \textbf{P}        & \textbf{R}        & \textbf{F1}       \\
        \hline
        NPN \cite{rashkin2018modeling}\rule{0pt}{2ex}
                            & 24.33             & 40.10             & 30.29             \\
        Llama3-8B-Instruct  & 36.20             & 35.51             & 35.23             \\
        \hline
        Ours\rule{0pt}{2ex}                & 43.55             & 42.68             & 42.98             \\
        Ours-\textit{affected}       & \textbf{48.51}    & \textbf{50.33}    & \textbf{49.03}    \\
        \hline\hline
    \end{tabularx}
    \caption{Result of Emotion Classification. The best performance is highlighted in bold, where 'Ours-\textit{affected}' means we consider the '\textit{affected}' features during classification, and the affective features denote whether a character is influenced by any emotion.}
    \label{tab:Result of Emotion Classification}
\end{table}

\section{Results and Analysis}
\subsection{Action Abstraction and Emotion Classification}
The action abstraction has summarized the key concept from the original sentence as open-text, so we evaluate it by creating a simple process called '\textit{Text to Action to Text (T2Act2T)}'. \textit{T2Act2T} takes the abstracted actions at first, and then it generates a new sentence only based on the abstracted actions. In the end, we compare the original sentence with the generated sentence to see how much information remained during the MLD-EA's action abstraction module.
Table \ref{tab:Result of Action Abstraction} shows the result between original sentence and new sentence, which illustrates the degree of information kept by our method.

We give results for emotion classification in Table \ref{tab:Result of Emotion Classification}. Our model performs best compared to Llama3-8B-Instruct baseline and a developed NPN model \cite{bosselut2017simulating} in ROCStories dataset \cite{rashkin2018modeling}. After fine-tuning, the model archives a significant improvement in emotion classification. Also, when incorporating the '\textit{affected}' feature to detect whether any emotion influences a character, our model attains an impressive F1 score of 88.51 on evaluating the accuracy of \textit{'affected'}, respectively. Our findings suggest that including features that account for emotional impact can dramatically improve classification performance, which has implications for various applications in natural language processing.

\begin{figure}[h]
  \includegraphics[width=\columnwidth]{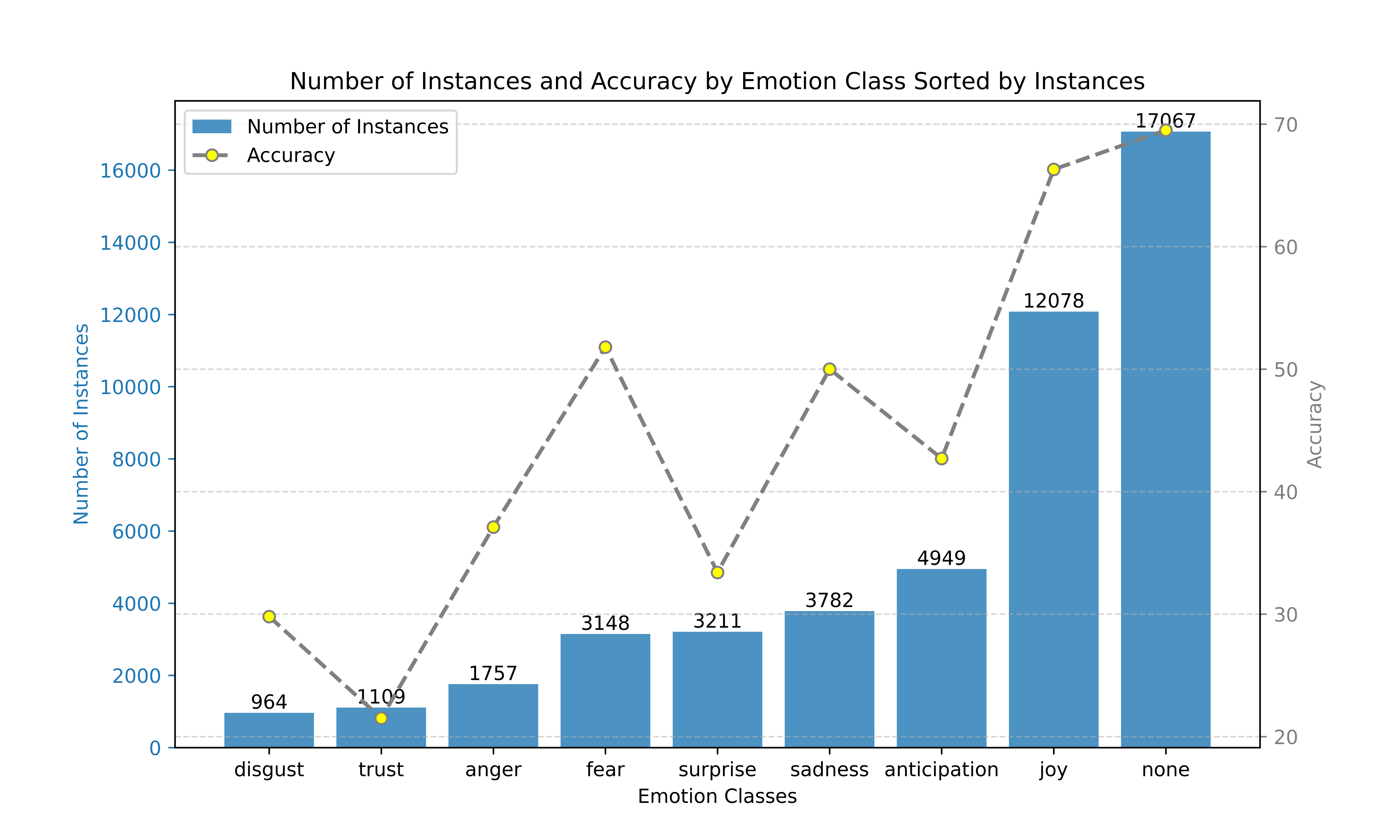}
  \caption{Emotion classification. This figure illustrates the relationship between the number of instances for each emotion class and the corresponding classification accuracy. Classes with more instances, such as 'joy,' exhibit higher classification accuracy compared to less frequent classes like 'disgust' and 'trust,' reflecting the potential influence of data imbalance on performance.}
  \label{fig:emotion_classification_chart}
\end{figure}

Furthermore, the partition relationship between the number of labels and their classified accuracy is in Figure \ref{fig:emotion_classification_chart}. Classes with fewer instances show lower accuracy, indicating a need for better representation or enhanced feature engineering to improve performance across less frequent emotions.

\begin{table*}[ht]
\centering
\scriptsize
  \begin{tabularx}{\linewidth}{p{2.25cm}YYYYYYYYYYYYYYY}
    \hline\hline \rule{0pt}{2ex}
    \textbf{Model}   & \multicolumn{3}{c}{\textbf{$k$=-1}}  & \multicolumn{3}{c}{\textbf{$k$=2}}  & \multicolumn{3}{c}{\textbf{$k$=3}}  & \multicolumn{3}{c}{\textbf{$k$=4}}  & \multicolumn{3}{c}{\textbf{Avg}} \\
          & \textbf{P} & \textbf{R} & \textbf{F1} & \textbf{P} & \textbf{R} & \textbf{F1} & \textbf{P} & \textbf{R} & \textbf{F1} & \textbf{P} & \textbf{R} & \textbf{F1} & \textbf{P} & \textbf{R} & \textbf{F1}                                                   \\
    \hline
    Without \textbf{EA}  \rule{0pt}{2ex}                                                                                                               \\
    \hline  
    Llama3-8B-Instruct\rule{0pt}{2ex}  & 0.00   & 0.00   & 0.00   & 17.90  & 64.43  & 24.80  & 89.72  & 29.47  & 44.29  & 14.04  & 57.82  & 19.47  & 30.41  & 37.93  & 22.14 \\
    Gemma2-2B-it       & 0.00   & 0.00   & 0.00   & 2.01   & 38.46  & 3.81   & 86.47  & 28.06  & 42.31  & 7.68   & 36.24  & 12.42  & 24.04  & 25.69  & 14.64\\
    Gemma2-9B-it       & 22.44  & 21.73  & 21.79  & 23.94  & 83.03  & 29.37  & 81.45  & 34.93  & 48.28  & 31.58  & 66.97  & 41.23  & 39.85  & 51.66  & 35.17\\ 
    \hline
    With \textbf{EA*}\rule{0pt}{2ex} \\
    \hline 
    Llama3-8B-Instruct*\rule{0pt}{2ex} & 0.64   & 11.11  & 1.21   & 32.44  & 48.25  & 27.86  & 56.39  & 32.11  & 40.75  & 39.47  & 53.98  & 35.27  & 32.24  & 36.11  & 26.27 \\
    Gemma2-2B-it*       & 15.38  & 10.74  & 9.53   & 60.18  & 34.29  & 43.67  & 29.32  & 27.67  & 26.88  & 6.36   & 60.13  & 10.56  & 27.81  & 33.21  & 22.66 \\
    Gemma2-9B-it*       & 29.49  & 37.83  & 28.62  & 24.61  & 54.55  & 30.51  & 73.18  & 34.58  & 46.36  & 38.60  & 67.50  & 48.09  & 41.47  & 48.62  & 38.39 \\
    \hline 
    MLD-EA (Ours)\rule{0pt}{2ex}      & 93.44  & 100.00  & 96.61  & 73.08  & 74.03  & 73.55  & 82.93  & 56.67  & 67.33  & 54.79  & 85.10  & 66.67  & \textbf{76.06}  & \textbf{78.95}  & \textbf{76.04} \\
    \hline\hline
  \end{tabularx}
  \caption{Result of Narrative Logic Checker on predicting missing sentence position. The best performance on average is highlighted in bold. \textbf{$k$=-1}: The input story is completed; \textbf{$k$=2,3,4}: The missing one should be inserted before index[2,3,4], where story's index starts at 1; \textbf{Avg}: the Micro-average score of all index's F1 score; Without \textbf{EA}: prediction without involving emotions and actions. With \textbf{EA} and \textbf{*}: prediction involving emotions and actions. }
  \label{tab:Result of Missing Sentence Prediction}
\end{table*}

\begin{table*} [ht]
\centering
\scriptsize
  \begin{tabularx}{\linewidth}{p{3.5cm} YYYYYYYYY}
    \hline\hline \rule{0pt}{2ex}
    \textbf{Model}             & \multicolumn{3}{m{2.15cm}}{\textbf{BLEU}}     & \multicolumn{3}{m{2.3cm}}{\textbf{ROUGE}}    & \multicolumn{3}{m{2.45cm}}{\textbf{BERTScore}} \\
         & \textbf{-1}   & \textbf{-2}  & \textbf{-4}  & \textbf{-1}  & \textbf{-2}  & \textbf{-L}  & \textbf{P}  & \textbf{R}  & \textbf{F1}\\
    \hline
    Without \textbf{EA} \rule{0pt}{2ex}  \\
    \hline
    Llama3-8B-Instruct\rule{0pt}{2ex}      & 33.77  & 4.05   & 0.28  & 25.98  & 5.56   & 22.27  & 77.66  & 79.03  & 78.33  \\
    Gemma2-2B-it            & 30.31  & 2.88   & 0.15  & 23.36  & 3.92   & 20.19  & 76.91  & 78.36  & 77.67 \\
    Gemma2-9B-it            & 33.82  & 3.38   & 0.14  & 24.42  & 4.03   & 20.84  & 77.76  & 78.94  & 78.34 \\
    \hline
    With \textbf{EA*} \rule{0pt}{2ex} \\
    \hline
    Llama3-8B-Instruct*\rule{0pt}{2ex}     & 36.29  & 5.83   & 0.54  & 28.18  & 7.18   & 23.99  & 77.59  & 78.98  & 78.27  \\
    $\text{Llama3-8B-Instruct}^\mp$*   & 43.68  & 12.15  & \textbf{2.67}  & 34.77  & 14.23  & 31.13  & 77.91  & 79.34  & \textbf{78.61} \\
    Gemma2-2B-it*           & 33.74  & 6.35   & 1.84  & 28.30  & 8.66   & 25.52  & 76.59  & 78.52  & 77.54 \\
    $\text{Gemma2-2B-it}^\mp$*         & 35.98  & 7.80   & 2.42  & 31.03  & 11.37  & 27.35  & 76.54  & 78.25  & 77.37 \\
    Gemma2-9B-it*           & 37.27  & 4.99   & 0.35  & 27.08  & 5.78   & 23.65  & \textbf{77.92}  & 78.90  & 78.40 \\
    $\text{Gemma2-9B-it}^\mp$*        & 40.14  & 7.83   & 0.91  & 30.56  & 9.01   & 26.82  & 77.87  & \textbf{79.21}  & 78.53 \\
    \hline
    \textbf{Pre-training Models }\\
    \hline
    COINS$^\dagger$ \cite{paul2021coins} & 22.82 & 10.52 & - & - & - & 19.4 & - & - & - \\
    CHAE$^\dagger$ \cite{wang2022chae} & 32.04 & \textbf{15.89} & - & - & - & - & - & - & - \\
    COG-BART$^\dagger$ \cite{xie2022comma} & 24.51 & 2.26 & 0.16 & 18.71 & 3.11 & 17.24 & - & - & - \\
    \hline
    MLD-EA (Ours)\rule{0pt}{2ex}       & \textbf{43.92}  & 12.17  & 2.29  & \textbf{35.51}  & \textbf{14.48}  & \textbf{31.41}  & 76.34  & 77.84  & 77.08 \\
    \hline
  \end{tabularx}
\caption{Result of Missing Sentence Generation. The best performance is highlighted in bold. \textbf{EA} and \textbf{*}: Emotions and Actions involved; $\mp$: Input with the action-emotion prediction of the missing sentence. $\dagger$: The results are taken from the highest scores from their research output.}
\label{tab:Result of Missing Sentence Generation}
\end{table*}

\subsection{Narrative Logic Checker}
Table \ref{tab:Result of Missing Sentence Prediction} presents the results of the narrative logic checker on predicting the index of missing part, which evaluates our model against various baselines both with and without incorporating actions and emotions\footnote{The full results of baselines are shown in Appendix \ref{Details of Narrative Logic Checker result on predicting missing sentence index}}. MLD-EA model consistently outperforms all baselines across different sentence insertion points. Notably, including actions and emotions significantly improves the micro-averaged F1 scores for all baseline models. Specifically, when the story is complete ($k=-1$), there is a marked improvement in F1 scores for each baseline model, underscoring the critical role that action and emotion play in maintaining story logic. 

The superior performance of our MLD-EA model highlights its advanced capability to accurately predict the missing sentence in a narrative. This suggests that the model's ability to consider emotional and action-related cues is essential for enhancing the logical coherence of stories. These findings emphasize the importance of incorporating nuanced narrative elements, such as emotions and actions, in developing more sophisticated and reliable models for story generation.

\subsection{Sentence Generation}
We also compare our model with the baselines for the Generation task, which considers the different situations. Also, we add the influence of action-emotion prediction on generation task. The results, as shown in Table \ref{tab:Result of Missing Sentence Generation}, demonstrate that our MLD-EA model, particularly when incorporating predicted actions and emotions, achieves competitive performance across multiple metrics. Notably, our model with the action-emotion prediction achieves the highest scores in several key areas: BLEU-1, BLEU-2, and all ROUGE. Moreover, we notice that BLEU-4 rises dramatically after involving emotions and actions for the Gemma2-2B-it model. This means this method may be more suitable for small-size LLMs on generation tasks with consistency and coherency.
We also compare this with previous studies in story plot generation, which are done by pre-training models. Obviously, the LLMs-based results achieve impressive improvement in generation tasks.

Incorporating actions and emotions into the generation process significantly enhances the model's performance, as evidenced by the notable improvement in BLEU and ROUGE scores across all baselines. 
However, the difference in BERTScore is slight. Overall, the baselines involved in emotion and action while adding action-emotion prediction still outperform the fundamental baselines.

We also use VAD to measure the deviation from the original sentence with model generation. Table \ref{tab:VAD} concludes that both LLMs can make the generated sentence closer to an original sentence in emotional dimensions after introducing emotions and actions. This improvement indicates that incorporating emotional and action cues enhances the logical consistency of the narrative and ensures that the generated content aligns more closely with the emotional tone of the original text, making the output more authentic and contextually appropriate.

\begin{table}[h]
\centering
\normalsize
\resizebox{\linewidth}{!}{
    \begin{tabular}{lcccccc}
    \hline\hline
    \textbf{Model}    & \textbf{V}    & \textbf{A}    & \textbf{D}    & \textbf{MEAN}     & \textbf{AoA}  & \textbf{Con}\\
    \hline
    Without \textbf{EA} \\
    \hline
    Llama3-8B-Instruct                  & 0.160 & 0.095 & 0.122 & 0.126 & 0.068 & 0.140 \\
    Gemma2-2B-it                        & 0.176 & 0.092 & 0.129 & 0.133 & 0.069 & 0.160 \\
    Gemma2-9B-it                        & 0.154 & 0.093 & 0.113 & 0.120 & 0.070 & 0.153 \\
    \hline
    With \textbf{EA*} \\
    \hline
    Llama3-8B-Instruct*                  & 0.157 & 0.092 & 0.123 & 0.124 & 0.064 & 0.143 \\
    Gemma2-2B-it*                        & 0.165 & 0.092 & 0.111 & 0.123 & \textbf{0.063} & 0.148 \\
    Gemma2-9B-it*                        & 0.143 & 0.095 & \textbf{0.110} & \textbf{0.116} & 0.066 & 0.154 \\
    \hline
    MLD-EA (Ours)                        & \textbf{0.142} & \textbf{0.092} & 0.116 & 0.117 & 0.065 & \textbf{0.137} \\
    \hline\hline
    \end{tabular}
}

\caption{VAD: deviation between original sentence and generated sentence. The closest result is highlighted in bold; \textbf{V}: Valence; \textbf{A}: Arousal; \textbf{D}: Dominance; \textbf{MEAN}: mean values of VAD; \textbf{AoA}: Age of Acquisition; \textbf{Con}: Concreteness; All values range from 0 to 1.}
\label{tab:VAD}
\end{table}
\begin{table}[hb]
    \centering
    \footnotesize
        \begin{tabularx}{\linewidth}{m{1.5cm}<{\centering}m{1.25cm}<{\centering}m{1.5cm}<{\centering}X}
        \hline\hline
        \textbf{Model}           & \textbf{P}        & \textbf{R}        & \textbf{ }\textbf{ }\textbf{ }\textbf{ }\textbf{ }\textbf{ }\textbf{ }\textbf{ }\textbf{ F1}   \rule{0pt}{2ex}\\
        \hline
        MLD-EA   & 81.09              & 81.19              & 80.89  \rule{0pt}{2ex}\\
        \textbf{w/o ae}         & 77.20           & 77.24               & 77.12 $\textcolor{red}{\downarrow2.97}$   \\
        \textbf{w/o a}          & 69.28             & 69.23             & 69.16 $\textcolor{red}{\downarrow10.93}$ \\
        \textbf{w/o e}          & 66.47             & 66.84             & 66.64 $\textcolor{red}{\downarrow13.45}$ \\
        \hline\hline
        \end{tabularx}
    \caption{Ablation Study of Narrative Logic Checker on predicting missing sentence position with conditional training. \textbf{w/o ae}: without actions and emotions; \textbf{w/o a}: without actions, emotions only; \textbf{w/o e}: without emotions, actions only.}
    \label{tab:Ablation Study}
\end{table}

\subsection{Ablation Study}

MLD-EA's primary task is to find the logic gap by providing characters' emotions and actions. So, we focus on how actions and emotions affect the model's performance in ablation studies. As shown in Table \ref{tab:Ablation Study}\footnote{Original emotions labels fine-tune all model here}, after we remove actions (\textbf{w/o a}) and emotions (\textbf{w/o e}), the accuracy of prediction drops out a lot, which decreases by more than 10 on micro-average F1 score. Surprisingly, we notice if we do not add actions and emotions as input (\textbf{w/o ae}), the prediction result is not bad. It may be because the default considering features is unrelated to characters' behaviors; it may be more related to the inherent connection between sentences.

\begin{table}[h]
\definecolor{f3e6f1}{RGB}{243,230,241}
\definecolor{fdcee4}{RGB}{253,206,228}
\definecolor{c77fa1}{RGB}{211,152,190}
    \centering

    \resizebox{\linewidth}{!}{
        \begin{tabular}{p{2cm}|p{8cm}}
        \hline
        Input Story:  &   $s_0$: "\textit{Gary was looking for a new laptop.}" 
                        \newline  $s_1$: "\textit{Gary needed the laptop to be light, fast, and efficient.}" 
                        \newline  $s_2$: "\textit{After researching, Gary decided he would purchase a Mac Air.}"
                        \newline  $s_3$: "\textit{Gary was not pleased with the operation of his new laptop at all.}" \\
                        
        \rowcolor{fdcee4}
        Character: &  Gary \\
        \rowcolor{c77fa1}
        Actions:  & LookingFor(a new laptop) $\rightarrow$ Needed(laptop) $\rightarrow$ Decided(Purchase a Mac Air) $\rightarrow$ NotPleased(laptop) \\
        \rowcolor{f3e6f1}
        Emotion:  & anticipation $\rightarrow$ anticipation $\rightarrow$ joy $\rightarrow$ anger \\
        \rowcolor{fdcee4}
        Missing Sentence Index: & Insert before sentence \textbf{3}. \\
        \rowcolor{c77fa1}
        Missing Sentence:   & "\textit{After purchasing, Gary quickly realized he made the wrong decision.}" \\
        \hline\hline
        Without \textbf{EA}  \\
        \hline
        \rowcolor{gray!20}
        Llama3-8B-Instruct  & Insert before: 2 \newline Sentence: "\textit{He had heard great reviews about the Mac Air's performance and durability.}"    \\
        \rowcolor{gray!40}
        Gemma2-2B-it        & Insert before: 2 \newline Sentence: "\textit{Despite its sleek design and powerful processor, the Mac Air seemed to be plagued by frequent crashes and slowdowns.}"    \\
        \rowcolor{gray!60}
        Gemma2-9B-it        & Insert before: 3 \newline Sentence: "\textit{Despite its sleek design and positive reviews, the Mac Air proved to be frustratingly slow.}"    \\
        \hline\hline
        With \textbf{EA}  \\
        \hline
        \rowcolor{gray!20}
        Llama3-8B-Instruct  & Insert before: 3 \newline Sentence: "\textit{Gary was frustrated with the slow performance and poor battery life of his new laptop.}"    \\
        \rowcolor{gray!40}
        Gemma2-2B-it        & Insert before: 2 \newline Sentence: "\textit{After reading reviews and comparing specs, Gary felt a surge of excitement at the prospect of finally getting a Mac Air.}"    \\
        \rowcolor{gray!60}
        Gemma2-9B-it        & Insert before: 3 \newline Sentence: "\textit{However, the Mac Air proved to be much slower than he had anticipated.}"      \\
        \hline
        \end{tabular}  
    }
    \caption{Example: different models with different rules on missing sentence index prediction and generation.}
    \label{tab:case study}
\end{table}

\subsection{Case study}

Table \ref{tab:case study} shows the result of models in finding the missing one's index and generating the sentence before and after the introduction of action and emotion. Without actions and emotions, most models incorrectly predicted the missing location, generating sentences that did not align with the emotional progression. For example, Llama3-8B-Instruct suggested inserting a sentence before $s_2$ that did not logically lead to Gary's later frustration.

When actions and emotions were included, model performance improved significantly. Both Llama3-8B-Instruct and Gemma2-9B-it accurately identified the correct index and generated sentences that better reflected the emotional shift from joy to anger, such as "\textit{Gary was frustrated with the slow performance and poor battery life of his new laptop.}". The example of "\textit{However, the Mac Air proved to be much slower than he had anticipated.}" even reflects the previous emotion status, making the sentence more connective to the story's consistent emotions and actions. This case study highlights the importance of action-emotion modeling in enhancing the accuracy and coherence of narrative generation, leading to more logically consistent and emotionally resonant outputs.

\section{Conclusion}
In this work, we introduced the MLD-EA model, a novel approach that leads LLMs to address gaps in narrative logic by integrating actions and emotions. MLD-EA extracts the actions and emotions of the characters in the input story and guides LLMs to find logical loopholes in the narrative by following the rules of interaction between actions and emotions. After getting the position where the missing part should be inserted, it combines the character behaviors and emotions in the context of the missing position to predict the possible character actions and emotions and complete the missing plot. The experimental results demonstrate that MLD-EA significantly improves narrative coherence and emotional alignment compared to existing models, highlighting its effectiveness in story logic detection and generation. By focusing on the interplay between actions and emotions, we have shown that maintaining logical consistency is crucial for producing believable and emotionally resonant narratives. This work advances the field of checking story logic and showcases the potential of LLMs as powerful tools for ensuring narrative cohesion.

\section*{Limitations}
First, the model has only been tested on short, five-sentence stories and has yet to be evaluated on longer, more complex narratives. This may limit its generalizability to extended storytelling contexts. Second, the model's performance heavily relies on the quality of the original emotion labels and action abstractions. Any inaccuracies in these inputs could negatively affect the model's ability to generate coherent and logically consistent narratives. Future work should address these limitations by testing the model on longer stories and improving the robustness of emotion and action extraction.

\section*{Acknowledgments}
This work is supported by the Alan Turning Institute/DSO grant: Improving multimodality misinformation detection with affective analysis. Yunfei Long, and Jinming Zhang acknowledge the financial support of the School of Computer Science and Electrical Engineering, University of Essex.

\bibliography{coling_latex}

\appendix

\section{Chosen Missing Sentence} \label{Chosen Missing Sentence}
Given a sequence of emotions attributed to characters in a narrative, we determine where emotional changes are most pronounced. Specifically, we analyze the emotions expressed by each character at different steps, calculate the "distance" between emotions in sentences, and identify the step where the aggregate emotional change across all characters is the greatest. This is crucial for understanding key moments in emotional narratives, potentially highlighting climaxes or critical turning points.

For each character $c$, at current sentence $s_i$, we calculate the emotion change value $D(e_{s_i}, e_{s_j}, c)$ for each sentence $s_j$:
\begin{equation}
    D(e_{s_i}, e_{s_j}, c) = \begin{cases} 
    \frac{d(e_{s_i}^c, e_{s_j}^c)}{|s_i - s_j|} & \textit{if }  e_{s_i,c} \\ 
                                                & \textit{and } e_{s_j,c} \\
    0                                           & \textit{otherwise}
\end{cases}
\end{equation}
where $d(x)$ represents the function to compute the distance between $e_{si}^c$ and $e_{s_j}^c$. Then, identify the sentence $s_{i_{max}}$ where emotions change maximized:
\begin{equation}
    s_{i_{max}} = \arg\max_{s_i}\sum_{c=c_1}^{c_m}\sum_{s_j=s_1}^{s_n} D(e_{s_i}, e_{s_j}, c), 
\end{equation}
where \(i_{\max}\) represents the index in the sequence where the emotions across all characters experience the greatest change. Then we remove this $s_{i_{max}}$ from the original story.

\section{hyper-parameters Used in MLD-EA} \label{hyper-parameters of fine-tune for MLD-EA}
Table \ref{tab:hyper-parameters of fine-tune} shows hyper-parameters of fine-tuning. The generation tasks' hyper-parameters for all models are the same as shown in Table \ref{tab:hyper-parameters of generation}.

\begin{table}[htbp]
    \centering
    \footnotesize
    \begin{tabularx}{\linewidth}{X|X}
    \textbf{Parameter name} & \textbf{Value} \\
    \hline
        \verb|lora_rank| \rule{0pt}{2ex}    & $8$   \\ \hline
        \verb|lora_alpha| \rule{0pt}{2ex}  & $16$  \\ \hline
        \verb|lora_dropout|\rule{0pt}{2ex} & $0.1$ \\ \hline
        \verb|lora_target|\rule{0pt}{2ex}  &   \texttt{all}    \\ \hline
        \verb|learing rate|\rule{0pt}{2ex} & $2e-5$ \\ \hline
        \verb|epoches|\rule{0pt}{2ex}      & $3$   \\
    \hline
    \end{tabularx}
    \caption{hyper-parameters of fine-tuning}
    \label{tab:hyper-parameters of fine-tune}
\end{table}


\begin{table}[htbp]
    \centering
    \footnotesize
    \begin{tabularx}{\linewidth}{X|X}
    \textbf{Parameter name} & \textbf{Value} \\
    \hline
        \verb|torch_dtype| \rule{0pt}{2ex}    & \texttt{torch.float16}   \\ \hline
        \verb|do_sample| \rule{0pt}{2ex}  & \texttt{True}  \\ \hline
        \verb|temperature|\rule{0pt}{2ex} & $0.1$ \\ \hline
        \verb|top_p|\rule{0pt}{2ex}  &   $0.4$   \\ 
    \hline
    \end{tabularx}
    \caption{hyper-parameters of generation}
    \label{tab:hyper-parameters of generation}
\end{table}

\begin{table*}[h]
\centering
\scriptsize
\begin{tabularx}{\linewidth}{p{1.25cm} | p{2.0cm} YYYYYYYYYYYYYYY}
    \hline\hline 
    \textbf{Actions and Emotions} & \textbf{Model}   & \multicolumn{3}{c}{\textbf{$k$=-1}}  & \multicolumn{3}{c}{\textbf{$k$=2}}  & \multicolumn{3}{c}{\textbf{$k$=3}}  & \multicolumn{3}{c}{\textbf{$k$=4}}  & \multicolumn{3}{c}{\textbf{Avg}} \\
    \textbf{} &\textbf{} & \textbf{P} & \textbf{R} & \textbf{F1} & \textbf{P} & \textbf{R} & \textbf{F1} & \textbf{P} & \textbf{R} & \textbf{F1} & \textbf{P} & \textbf{R} & \textbf{F1} & \textbf{P} & \textbf{R} & \textbf{F1}  \\
    \hline
    \multirow{12}{*}{\hfil Without \textbf{EA}} 
    &\textbf{Llama3-8B-Instruct}  \\
    &zero-shot      & 0.00  & 0.00  & 0.00  & 10.73  & 57.14  & 18.08  & 95.49  & 27.97  & 43.27  & 0.00   & 0.00   & 0.00   & 26.56  & 21.28  & 15.34 \\
    &one-shot       & 0.00  & 0.00  & 0.00  & 35.57  & 53.53  & 42.74  & 78.95  & 27.70  & 41.02  & 2.63   & 50.00  & 5.00   & 29.29  & 32.81  & 22.19 \\
    &few-shot       & 0.00  & 0.00  & 0.00  & 12.75  & 73.08  & 21.71  & 90.98  & 32.27  & 47.64  & 36.18  & 67.90  & 47.21  & 34.98  & 43.31  & 29.14 \\

    &\textbf{Gemma2-2B-it}  \\
    &zero-shot      & 0.00  & 0.00  & 0.00  & 2.68   & 57.14  & 5.13   & 98.50  & 28.42  & 44.10  & 5.26   & 44.44  & 9.41   & 26.61  & 32.50  & 14.66 \\
    &one-shot       & 0.00  & 0.00  & 0.00  & 1.34   & 15.38  & 2.47   & 76.69  & 26.98  & 39.92  & 0.00   & 0.00   & 0.00   & 19.51  & 10.59  & 10.60 \\
    &few-shot       & 0.00  & 0.00  & 0.00  & 2.01   & 42.86  & 3.85   & 84.21  & 28.79  & 42.91  & 17.76  & 64.29  & 27.84  & 26.00  & 33.98  & 18.65 \\

    &\textbf{Gemma2-9B-it} \\
    &zero-shot      & 0.00  & 0.00  & 0.00  & 3.36   & 1.00   & 6.49   & 95.49  & 29.74  & 45.36  & 23.68  & 67.92  & 35.12  & 30.62  & 49.42  & 21.74 \\
    &one-shot       & 32.69 & 24.29 & 27.87 & 55.03  & 62.12  & 58.36  & 68.42  & 38.89  & 49.59  & 23.68  & 73.47  & 35.82  & 44.96  & 49.69  & 42.91 \\
    &few-shot       & 34.62 & 40.91 & 37.50 & 13.42  & 86.96  & 23.26  & 80.45  & 36.15  & 49.88  & 47.37  & 59.50  & 52.75  & 43.96  & 55.88  & 40.85 \\ 
    \hline
    
    \multirow{12}{*}{With \textbf{EA*}}
    &\textbf{Llama3-8B-Instruct*} \\
    &zero-shot       & 0.00  & 0.00  & 0.00  & 77.86  & 39.46  & 52.37  & 36.09  & 25.40  & 29.81  & 1.32   & 66.67  & 2.58   & 28.81  & 32.88  & 21.19 \\
    &one-shot        & 0.00  & 0.00  & 0.00  & 14.09  & 43.75  & 21.32  & 57.89  & 33.19  & 42.19  & 52.63  & 41.24  & 46.24  & 31.16  & 29.54  & 27.44 \\
    &few-shot        & 1.92  & 33.33 & 3.64  & 5.37   & 61.54  & 9.88   & 75.19  & 37.74  & 50.25  & 64.47  & 51.04  & 56.98  & 36.74  & 45.91  & 30.19 \\

    &\textbf{Gemma2-2B-it*} \\
    &zero-shot       & 0.00  & 0.00  & 0.00  & 65.77  & 35.90  & 46.45  & 36.84  & 23.90  & 28.99  & 3.29   & 71.43  & 6.29   & 26.48  & 32.81  & 20.43 \\
    &one-shot        & 38.46 & 11.17 & 17.32 & 46.98  & 28.57  & 35.53  & 12.03  & 27.59  & 16.75  & 0.66   & 50.00  & 1.30   & 24.53  & 29.33  & 17.73 \\
    &few-shot        & 7.69  & 21.05 & 11.27 & 67.79  & 38.40  & 49.03  & 39.10  & 31.52  & 34.90  & 15.13  & 58.97  & 24.08  & 32.43  & 37.49  & 29.82 \\

    &\textbf{Gemma2-9B-it*} \\
    &zero-shot       & 1.92  & 33.33 & 3.64  & 12.75  & 57.58  & 20.88  & 90.23  & 30.77  & 45.89  & 26.32  & 66.67  & 37.74  & 32.80  & 47.09  & 27.04 \\
    &one-shot        & 38.46 & 44.44 & 41.24 & 48.32  & 51.80  & 50.00  & 61.65  & 37.10  & 46.33  & 36.18  & 70.51  & 47.83  & 46.16  & 50.96  & 46.35 \\
    &few-shot        & 48.08 & 35.71 & 40.98 & 12.75  & 54.29  & 20.65  & 67.67  & 35.86  & 46.88  & 53.29  & 65.32  & 58.70  & 45.45  & 47.79  & 41.80 \\
    \hline\hline
\end{tabularx}

\caption{Details of Narrative Logic Checker result on predicting missing sentence index. \textbf{-1}: The input story is completed; \textbf{2,3,4}: The missing sentence should be inserted before index[2,3,4], where story's index starts at 1; \textbf{Avg}: the Micro-average score of all index's F1 score; Without \textbf{EA}: prediction without involving emotions and actions. With \textbf{EA} and \textbf{*}: prediction involving emotions and actions.}
\label{tab:Details of Narrative Logic Checker result}
\end{table*}

\section{Details of Narrative Logic Checker result on predicting missing sentence index} \label{Details of Narrative Logic Checker result on predicting missing sentence index}
Table \ref{tab:Details of Narrative Logic Checker result} shows all results of the narrative logic checker running on baselines with different prompt techniques we used in the experiment.

All results of missing sentence index prediction results when involved actions and emotions have increased on average.
Especially before involving actions and emotions in inference, they are hard to recognize when the story is completed. However, after we add actions and emotions during inferring, the LLMs can recognize the completed story even with the zero-shot prompt (\textbf{Gemma2-9B-it*}). These results illustrate that considering the interaction between actions and emotions can extraordinarily improve LLMs' narrative logic checking.

\section{Error Analysis: Generation results with correct index} \label{Error Analysis: Generation results with correct index}
One key area for error analysis involves evaluating how well the model predicts the correct index for the missing sentence. Misplacement of the generated sentence can disrupt the logical flow of the narrative. Table \ref{tab:Generation results with correct index}  shows the generation results when the input of the missing sentence index is correct. In this evaluation, we focused on how predicted action-emotion affects the generation quality of the missing part. So, we will only consider when the index is predicted correctly by the narrative logic checker in relation to the analysis results, which involve emotion and actions. 

The results show the importance of when models predict the index of logical loopholes. The change of BLEU and ROUGE remains the same because they are all compared with the reference story. At the sentence level BERTScore measures, the F1 score increases dramatically if the generated sentence is filled in the right place. This highlights the model's ability to produce more contextually appropriate and coherent content when the narrative gap is accurately identified. This underscores the importance of accurate index prediction in generating logically and emotionally consistent stories.

\begin{table*}[h]
    \centering
    \scriptsize
    \begin{tabularx}{\linewidth}{p{3cm} YYYYYYYYY}
        \hline\hline \rule{0pt}{2ex}
         \textbf{Model}             & \multicolumn{3}{c}{\textbf{BLEU}}     & \multicolumn{3}{c}{\textbf{ROUGE}}    & \multicolumn{3}{c}{\textbf{BERTScore}} \\
             & \textbf{-1}   & \textbf{-2}  & \textbf{-4}  & \textbf{-1}  & \textbf{-2}  & \textbf{-L}  & \textbf{P}  & \textbf{R}  & \textbf{F1}\\
        \hline
        Llama3-8B-Instruct\rule{0pt}{2ex}      & 43.68  & 12.15  & 2.67  & 34.77  & 14.23  & 31.13  & 77.91  & 79.34  & 78.61  \\
        Llama3-8B-Instruct*   & 44.26  & 12.56  & 2.93  & 35.08  & 14.58  & 31.46  & 87.35  & 88.91  & 88.11 \\
        Gemma2-2B-it           & 35.98  & 7.80   & 2.42  & 31.03  & 11.37  & 27.35  & 76.54  & 78.25  & 77.37 \\
        Gemma2-2B-it*         & 36.42  & 6.96   & 1.55  & 30.34  & 10.15  & 27.08  & 80.81  & 81.95  & 81.04 \\
        Gemma2-9B-it           & 40.14  & 7.83   & 0.91  & 30.56  & 9.01   & 26.82  & 77.87  & 79.21  & 78.53 \\
        Gemma2-9B-it*       & 40.63 & 8.04   & 0.84  & 30.50  & 9.31   & 26.88  & 87.01  & 88.53  & 87.75 \\

        \hline\hline    

    \end{tabularx}
    \caption{Generation results with correct index. The model with \textbf{*}: Input with the correct prediction of the missing sentence index. All results are based on the correct missing sentence index as input.}
    \label{tab:Generation results with correct index}
\end{table*}

\section{Prompt Engineering} \label{Prompts Engineering}
We started at zero-shot for all the cases, then developed one-shot and few-shots after confirming the zero-shot prompt template. Also, we used Chain-of-Thought as an assistant prompt strategy.

For emotion classification, we begin our approach by deploying a suite of meticulously designed prompts to leverage the MLD-EA's capabilities in emotion classification, guiding the model to accurately discern and categorize the emotional spectrum associated with each character in a given sentence. After establishing a baseline performance using the inherent strengths of LLMs, we refine our MLD-EA model through a process inspired by the baseline. This refinement is achieved using supervised fine-tuning with a custom-tailored prompt that enhances the model's ability to detect and classify emotions more precisely for individual characters. This targeted fine-tuning boosts the model's proficiency, enhancing its analytical and emotional sentiment analysis capabilities.

There are some examples of prompt templates used for baselines on experiments. Table \ref{tab:Prompt template: Action Abstraction} shows the prompt template for action abstraction. We also present the prompt templates for both '\textit{Without \textbf{EA}}' and '\textit{With \textbf{EA}}' for the narrative logic checker and generation tasks. The prompt templates of '\textit{Without \textbf{EA}}' mean the LLMs need to find the logic loopholes and complete the plot only along with the input story, which the zero-shot prompt template for the narrative logic checker is shown in Table \ref{tab:Prompt template: Narrative Logic Checker} and the generation template is in Table \ref{tab:Prompt template: Sentence Generation}. The prompt templates of '\textit{With \textbf{EA}}' means the LLMs have to consider the characters' emotions and actions during those tasks, which the zero-shot prompt template for the narrative logic checker is shown in Table \ref{tab:Prompt template: Narrative Logic Checker} and the generation template is in Table \ref{tab:Prompt template: Sentence Generation}. Also, Table \ref{tab:Prompt template: Actions and Emotions Prediction} shows how we predict the actions and emotions for the missing part.
Notably, All the results of '\textit{Without \textbf{EA}}' are actually how LLMs face those tasks without any further information. Our study considers the interaction of actions and emotions to increase overall performances for LLMs on those tasks.

\begin{table*}[htbp]
    \centering
    \footnotesize
    \begin{tabularx}{\linewidth}{X}
    \hline
    \textbf{Instruction:}\rule{0pt}{2ex} \\
    You are an AI designed to abstract and categorize actions from given sentences. You will receive a sentence with a list of characters(The characters may or may not appear in the sentence but appear in the completed story; some Pronouns like He, she, etc. mean one of the characters provided in Input. Do not care about those characters who are not provided). \newline
    Your task is to identify any actions these characters perform and abstract them from the sentence in a specific format.\newline    
    The whole story of this sentence will be provided before the sentence to help you do the mentioned detection.\newline                            
    Format for abstraction:\newline                            
    For each character, specify the action they performed and the target of the action (if any) in the form <Character>Action(Target, ActionObject)</Character>.\newline
    (\newline
    Character: The character performing the action (i.e. Lucy, I, Lucy's mom, etc.).\newline
    Action: The action performed by the character (i.e. Love, Loved, Loves, See, Saw, Attack, Attacks, Attacked, Move, Moves, Moved, Move to, Come, Came, etc.). \newline
    Target: The target of the action (who or what the action is directed towards) (i.e. A Love B -> <A>Love(B)</A>).\newline
    ActionObject: The specific object related to the action (if any) (i.e. A give b an apply -> <a>Give(b, an apply)</a>).\newline
    )\newline
    If a character does not perform any action, output <Character>None</Character>.\\
    \hline
    \end{tabularx}
    \caption{Prompt template: Action Abstraction}
    \label{tab:Prompt template: Action Abstraction}
\end{table*}

\begin{table*}[htbp]
    \centering
    \footnotesize
    \begin{tabularx}{\linewidth}{X}
    \hline
    Without \textbf{EA}\rule{0pt}{2ex} \\
    \hline
    \textbf{Instruction:}\rule{0pt}{2ex} \\
    You are an AI assistant designed to analyze and evaluate user inputs for completeness and coherence. Your primary task is to determine whether the provided sequence of sentences is missing a sentence. If you think a sentence is missing, identify where the missing one should be inserted, i.e. if a sentence is missing between sentence 1 and sentence 2, the result should be inserted at index 2; otherwise, if you think no sentence is missing here, just output -1.\newline
    UserInput will provide a story with several sentences;\newline

    Note that the first and last sentences are constantly provided at the story's start and end; they should not be considered missing. Please find out the index of where the missing one should be inserted before which sentence.
    Only give the final output, and in this format: Insert before sentence [**i**]. \\
    \hline\hline
    With \textbf{EA}\rule{0pt}{2ex} \\
    \hline
    \textbf{Instruction:}\rule{0pt}{2ex}\\
You are an AI assistant designed to analyze and evaluate user inputs for completeness and coherence. Your primary task is to determine whether the provided sequence of sentences is missing a sentence. If you think a sentence is missing, identify where the missing one should be inserted, i.e. if a sentence is missing between sentence 1 and sentence 2, the result should be inserted at index 2; otherwise, if you think no sentence is missing here, just output -1. \newline
UserInput will provide a story with several sentences; characters' actions and emotions in sentences are shown after each sentence.\newline

Consider the following Rules while analyzing:\newline
**Rules**:\newline
- Emotion affects Action: Actions are often taken because of an emotion.\newline
- Action affects Emotion: Emotions can change due to actions taken.\newline
- Emotion and Action at the same time: Some actions demonstrate current emotions.\newline
- Consider the relationship between the emotions and actions of each character linked between sentences to identify any missing sentence.\newline
- Analyze each character's action and emotion chain to find the missing parts. \newline

Use the given rules and provided data to ensure a logical flow of events and completeness in the narrative. Note that the first and last sentences are constantly provided at the story's start and end; they should not be considered missing. Please find out the index of where the missing one should be inserted before which sentence.
Only give the final output, and in this format: Insert before sentence [**i**].\\
    \hline
    \end{tabularx}
    \caption{Prompt template: Narrative Logic Checker}
    \label{tab:Prompt template: Narrative Logic Checker}
\end{table*}

\begin{table*}[htbp]
    \centering
    \footnotesize
    \begin{tabularx}{\linewidth}{X}
    \hline
    Without \textbf{EA}: \rule{0pt}{2ex}\\
    \hline
    \textbf{Instruction:} \rule{0pt}{2ex}\\
    You are an AI assistant (Master in story writing) designed to help users analyze, evaluate, and complete stories by checking their completeness and coherence.\newline
    Generate a sentence to fill a gap in a narrative based on the surrounding context, ensuring the story remains coherent and complete.\newline

**Generate the Missing Sentence**:\newline
--Create a sentence that naturally fits into the narrative at the specified index.\newline
--Ensure the new sentence connects logically with the sentences before and after it, maintaining a smooth and coherent flow.\newline
--Match the style and tone of the existing story.\newline

UserInput will provide a story with several sentences, and the index of missing one should be inserted before.\\
    \hline\hline
    With \textbf{EA} but no prediction actions and emotions: \rule{0pt}{2ex}\\
    \hline
    \textbf{Instruction:} \rule{0pt}{2ex}\\
    You are an AI assistant (Master in story writing) designed to help users analyze, evaluate, and complete stories by checking their completeness and coherence.\newline
    Generate a sentence to fill a gap in a narrative based on the surrounding context, ensuring the story remains coherent and complete.\newline

**Generate the Missing Sentence**:\newline
--Create a sentence that naturally fits into the narrative at the specified index.\newline
--Ensure the new sentence connects logically with the sentences before and after it, maintaining a smooth and coherent flow.\newline
--Match the style and tone of the existing story.\newline

UserInput will provide a story with several sentences; characters' actions and emotions in sentences are shown after each sentence. The Characters in story and the index of the missing sentence should be inserted before.\\
    \hline\hline
    With \textbf{EA} and prediction actions and emotions: \rule{0pt}{2ex}\\
    \hline
    You are an AI assistant (Master in story writing) designed to help users analyze, evaluate, and complete stories by checking their completeness and coherence.\newline
    Generate a sentence to fill a gap in a narrative based on the surrounding context, ensuring the story remains coherent and complete.\newline

**Generate the Missing Sentence**:\newline
--Create a sentence that naturally fits into the narrative at the specified index.\newline
--Ensure the new sentence connects logically with the sentences before and after it, maintaining a smooth and coherent flow.\newline
--Match the style and tone of the existing story.\newline
--Consider whether the given actions and emotions are reasonable in this situation. Then, generate the sentence.\newline

**Notes**:\newline
1. The action form looks like this: Action(Target, ActionObject), where \newline
(Action: The action performed by the character (i.e. Love, Loved, Loves, See, Saw, Attack, Attacks, Attacked, Move, Moves, Moved, Move to, Come, Came, etc.).\newline
Target: The target of the action (who or what the action is directed towards) (i.e. A Love B -> A Love(B)).\newline
ActionObject: The specific object related to the action (if any) (i.e. A give b an apply -> A Give(b, an apply)).\newline
)\newline
2. The emotions are ONLY from Plutchik's eight basic emotions (joy, trust, fear, surprise, sadness, disgust, anger, anticipation) for the characters based on their likely emotional state based on the context and characters' actions. If 'none' means the characters do not have a discernible emotion or will not appear at this point.\newline

UserInput will provide a story with several sentences, and the index of missing one should be inserted before. Also, the predicted actions and emotions of characters that may happen in this missing sentence will be given.\\
    \hline 
    \end{tabularx}
    \caption{Prompt template: Sentence Generation}
    \label{tab:Prompt template: Sentence Generation}
\end{table*}

\begin{table*}[htbp]
    \centering
    \footnotesize
    \begin{tabularx}{\linewidth}{X}
    \hline
    \textbf{Instruction:} \rule{0pt}{2ex}\\
    You are an AI assistant (Master in story writing) designed to help users analyze, evaluate and complete stories by checking their completeness and coherence. Especially is good at action analysis and Plutchik's emotion analysis. \newline
    **Purpose**: \newline
Predict the most likely actions and emotions of characters for a sentence that should be inserted before the specified index in a story, ensuring the narrative remains coherent and logically connected. UserInput will provide a story with several sentences, all characters in the story and the index of missing sentences should be inserted. Think it step by step. \newline

**Contextual Analysis**:\newline
1. Examine the provided story and identify the events leading to the specified index; if the index is -1, no missing sentence needs to be generated here; stop responding and give 'none'.\newline
2. Focus on the actions and emotions of the characters in the story to understand their progression.\newline

**Action Prediction**:\newline
1. Predict the most likely action that would occur before the specified index. This prediction should be based on strong evidence from the surrounding context and reflect a logical progression in the narrative.\newline
2. The action should be in the open-text format and reflect what the character would logically do next based on previous actions, emotions and the situation.\newline
3. The action form looks like this: Action(Target, ActionObject), where 
(Action: The action performed by the character (i.e. Love, Loved, Loves, See, Saw, Attack, Attacks, Attacked, Move, Moves, Moved, Move to, Come, Came, etc.).\newline
Target: The target of the action (who or what the action is directed towards) (i.e. A Love B -> A Love(B)).\newline
ActionObject: The specific object related to the action (if any) (i.e. A give b an apply -> A Give(b, an apply)).\newline
)\newline

**Emotion Prediction**:\newline
1. Assign an emotion ONLY from Plutchik's eight basic emotions (joy, trust, fear, surprise, sadness, disgust, anger, anticipation) to the characters based on their likely emotional state based on the context and characters' actions.\newline
2. If the characters do not have a discernible emotion or will not appear at this point, use 'none'.\newline

**Reasoning**:\newline
1. Provide the predicted action and emotion for each character(s) that should appear in the missing sentence before the specified index.\newline
2. Ensure that the predicted actions and emotions consistently follow logical flow.\\
    \hline    
    \end{tabularx}
    \caption{Prompt template: Actions and Emotions Prediction}
    \label{tab:Prompt template: Actions and Emotions Prediction}
\end{table*}

\end{document}